\documentclass[11pt]{article}
\usepackage[margin=1in]{geometry}
\usepackage{amsmath,amssymb}
\usepackage{booktabs}
\usepackage{graphicx}
\usepackage{xcolor}
\usepackage[hidelinks]{hyperref}
\usepackage{natbib}
\usepackage{microtype}

\newcommand{\bpc}{\mathrm{bpc}}

\title{Reaching Every Position Without Searching:\\
Rotating Sparse Wiring on the Hypercube as a Substitute for Attention}
\author{Yoshiaki Takashita\thanks{This work was carried out independently and is not affiliated with any laboratory.}\\
  School of Law, Waseda University\\
  takashita@moegi.waseda.jp}
\date{September 2026}

\begin{document}
\maketitle

\begin{abstract}
Attention pays, at every layer and for every input, the cost of \emph{searching} for whom to connect.
We ask how far one can get with wiring that is fixed, sparse, and simply \emph{rotated} from layer to layer.
Treating the $n$ positions of a sequence as the vertices of a $\log_2 n$-dimensional hypercube and connecting each position, at layer $\ell$, to its neighbour along dimension $\ell \bmod \log_2 n$, information from every position reaches every other in $\log_2 n$ layers with $2n$ links per layer instead of $n^2$.
On a synthetic task that is unsolvable unless all positions are reached, this rotation matches all-to-all wiring at $1/32$ of the links, while the same sparse pattern held fixed across layers fails; what matters is that every dimension is touched, not the order.
On character-level language modelling of a public corpus (the first $12$M characters of enwik8), a hybrid that keeps two attention layers among sixteen sparse ones reaches $0.06$ bits-per-character \emph{lower} held-out loss than a fully attentive model of the same width at the same step budget (three seeds each, no overlap), with $1/7$ of the links, $42\%$ fewer parameters, and $2.4\times$ less wall-clock time; the purely rotated schedule is level with the hybrid.
The same ordering holds on a second corpus of mixed Japanese, English and code, where the gap widens to $0.16$.
The usable learning-rate window is four to eight times wider than attention's on both.
We also report what did not work---learned coordinates, and a ``dynamics'' variant whose apparent gains turned out to be an artefact of a saturated kernel---and the measurement discipline (frozen corpus, full-coverage evaluation, seed spread as the bar for ranking) that we found necessary to say anything at all at this scale.
\end{abstract}

\section{Introduction}
\label{sec:intro}

The expensive part of attention is not the mixing; it is the search.
Every layer, for every input, attention recomputes which positions should influence which, and pays $O(n^2)$ for the privilege.
The question in this paper is deliberately narrow: if the wiring is \emph{fixed}---chosen once, independent of the input---how much of attention's quality survives, and at what fraction of its cost?

Fixed sparse patterns are not new \citep{child2019sparse,beltagy2020longformer,zaheer2020bigbird}; nor is the observation that a butterfly or hypercube schedule touches every position in $\log n$ stages \citep{dao2019butterfly}.
Our contribution is to isolate one variable---\emph{rotation}: changing which hypercube dimension each layer uses---and to measure it against controls that hold everything else constant (same weights, same data, same steps, same number of links), first on a synthetic task where reachability is the whole problem, then on language.
Along the way we found that several of our own intermediate ``results'' were measurement artefacts, and we report those as findings too (\S\ref{sec:negative}).

\paragraph{Claims.}
(A) Rotation is what makes sparse wiring reach: with the same $2n$ links per layer, rotating the dimension matches all-to-all wiring and the un-rotated control does not (\S\ref{sec:synthetic}).
(B) On language, at window $256$, a hybrid with two attention layers among sixteen sits $0.06\,\bpc$ below full attention on a public corpus and $0.16\,\bpc$ below it on a private one, at the same step budget, with $1/7$ of the links and a usable learning-rate window four to eight times wider (\S\ref{sec:language}).  At window $64$ the hybrid only draws level with attention, and the purely sparse schedules fall behind dense mixing---a pre-freeze claim of ours that did not survive remeasurement (\S\ref{sec:negative}).
(C) A hierarchy in which a coarse hypercube gates the fine one changes \emph{trainability}, not capacity: same asymptote, $30$ points faster at $500$ steps, $43\times$ smaller seed spread (\S\ref{sec:hierarchy}).
(D) A ``dynamics'' variant, in which positions move under learned forces, produced its early gains through a saturated kernel that reduced the model to a pointer chain; after fixing the scale the sign of the effect of the number of steps reverses (\S\ref{sec:dynamics}).

\section{Related work}
\label{sec:related}

Sparse attention keeps the softmax search but restricts where it may look: strided and local patterns \citep{child2019sparse}, a sliding window with a few global tokens \citep{beltagy2020longformer}, or random links added to both \citep{zaheer2020bigbird}.  Routing methods go the other way and learn the pattern per input, by hashing \citep{kitaev2020reformer} or by clustering \citep{roy2021routing}.  In both families the wiring is still decided at run time, which is the cost we set out to remove.

A second family removes the search altogether.  MLP-Mixer mixes tokens with a fixed dense map over positions \citep{tolstikhin2021mixer}; Hyena replaces attention with long convolutions whose filters are fixed once trained \citep{poli2023hyena}; Mamba keeps a selective, input-dependent state-space recurrence \citep{gu2023mamba}.  These operators are dense in position or carry per-position state.  Ours is a single shift and two linear maps per layer, with $2n$ links.

The structure we use is old.  Hypercube and butterfly networks reach every node in $\log n$ stages and are standard in parallel computing \citep{leighton1992parallel}; butterfly factorisations have been used to learn fast linear transforms \citep{dao2019butterfly}.  We do not claim the structure.  What we isolate is one variable, rotating the dimension from layer to layer, and we measure it against a control that keeps the same links but does not rotate.

\section{Method}
\label{sec:method}

\subsection{Positions as hypercube vertices}
Let $n = 2^b$.  We index positions by $b$-bit integers and, at layer $\ell$, connect position $i$ to $i \oplus 2^{(\ell \bmod b)}$---its neighbour along one hypercube dimension.
Each layer therefore has $2n$ directed links (plus the identity).  After $b$ consecutive layers every pair of positions is connected by a path; after $2b$ layers, in practice, the network uses the connectivity (\S\ref{sec:synthetic}).
For causal language modelling we keep only the earlier of the two endpoints, which turns the schedule into a parallel prefix scan.

\subsection{Mixing on a fixed link}
A sparse layer with jump $k$ computes
\begin{equation}
  y_i \;=\; W_{\mathrm{self}}\, x_i \;+\; W_{\mathrm{off}}\, x_{i-k},
  \label{eq:mixa}
\end{equation}
two linear maps and a shift.  There is no input-dependent routing.  This is the only operation the sparse layers use, and it is why the models export to ONNX with standard operators alone (\S\ref{sec:deploy}).

\subsection{Schedules}
We compare, holding weights, data, optimiser and steps fixed:
\begin{itemize}
  \item \textbf{dense}: all-to-all mixing ($n^2$ links per layer);
  \item \textbf{local}: $k=1$ at every layer;
  \item \textbf{fixed}: one hypercube dimension, the same at every layer (the control for ``sparse but not rotated'');
  \item \textbf{rot}: hypercube dimension $\ell \bmod b$ at layer $\ell$;
  \item \textbf{mixa}: local and long-range layers alternate, the long jumps visiting each scale once ($32,16,8,4,2,1$ for $n=64$);
  \item \textbf{hybrid}: the \textbf{mixa} backbone with two of the layers replaced by standard multi-head attention.
\end{itemize}
The exact layer lists are given as short schedule strings in Appendix~\ref{app:glyph}.

\subsection{Hierarchy and dynamics}
\label{sec:method-dyn}
Two further variants are evaluated on synthetic tasks.
In the \emph{hierarchical} variant, positions are grouped; each group is a vertex of a coarser hypercube that moves under the same rule and gates the couplings inside its group.  The force law is shared between levels, so the hierarchy adds $225$ parameters ($+1.8\%$) to a $12{,}482$-parameter model.
In the \emph{dynamics} variant, each position carries a coordinate $p_i \in \mathbb{R}^d$; couplings are $\exp(-\|p_i - p_j\|^2 / \tau)$, a learned force moves the coordinates, and this is repeated $T$ times before reading out.  The scaled form (``field'') divides the squared distance by its causal row mean before the exponential; the ``phase'' form additionally carries a velocity.

\section{Synthetic: rotation is what reaches}
\label{sec:synthetic}

\paragraph{Task.}  Count the ones in a random binary sequence of length $n$, reading the answer from position $0$ only.  The task cannot be solved unless information from every position reaches position $0$.  Guessing the most common answer gives $9.9\%$ for $n=64$.

\paragraph{Result (Claim A).}  With $n=64$, six layers, and identical weights and link counts for the two sparse schedules:
\begin{center}
\begin{tabular}{lrr}
\toprule
schedule & accuracy & links / layer \\
\midrule
fixed (not rotated) & 10.2\% & 768 \\
\textbf{rot} & \textbf{98.7\%} & \textbf{768} \\
dense & 100.0\% & 24{,}576 \\
\bottomrule
\end{tabular}
\end{center}
Sparsity is not what helps; rotation is.  The un-rotated control sits at chance with exactly the same number of links.

\paragraph{Depth.}  Accuracy jumps from $14.2\%$ at five layers to $98.8\%$ at six.  Six is $\log_2 64$, the depth at which the first complete path exists.  It saturates near $2\log_2 n$, with $99.8\%$ at $12$ layers and $100\%$ at $16$.  For $n=256$ the floor $\log_2 n = 8$ gives $28.0\%$ and $16$ layers give $52.1\%$.  The effective depth is again about twice the floor.

\paragraph{Order does not matter; coverage does.}  A random permutation of dimensions reaches $99.0\%$.  A Gray-code order, which visits dimension $0$ eight times and dimension $5$ never within $16$ layers, reaches $14.4\%$.  Every dimension has to be touched.  The order in which they are touched does not matter.

\section{Language}
\label{sec:language}

\paragraph{Setup.}  Character-level next-character prediction, reported as held-out bits per character.  Evaluation covers the validation set in non-overlapping windows, so the measurement noise is zero; what remains is optimiser-trajectory variance, which we estimate with seeds.  Learning rate is swept on a grid for \emph{every} schedule---early in this work we tuned it for one side only and drew the wrong conclusion three times (\S\ref{sec:negative}).

We measure on \emph{two} corpora, each frozen and identified by a fingerprint of its contents.  The first is public: the first $12$M characters of enwik8, rebuilt from the published archive by checksum (fingerprint \texttt{c52380b41455}, $201$ symbols).  The second is $12$M characters of mixed Japanese, English and code taken from this project's repository at a pinned commit (fingerprint \texttt{e1634705154f}, $4{,}769$ symbols).  The public corpus is the one a reader can reproduce; the second is kept because it is a different distribution, and the point of interest is whether the ordering survives the change.  We do \emph{not} compare our enwik8 numbers with the published literature: we use an $8\%$ block split rather than the customary $90/5/5$, so the numbers are internally comparable and externally not.

\subsection{Public corpus, window 256, 20--35M parameters}
\label{sec:w256e}
Sixteen layers, width $512$, $3{,}000$ steps.  The learning rate is swept over seven values and the best setting is repeated with three seeds.  At its best rate, attention reaches $1.778\,\bpc$ (seeds $1.772$, $1.771$, $1.791$) with $526{,}336$ links and $35.0$M parameters.  The \textbf{hybrid} reaches $1.718$ ($1.703$, $1.712$, $1.740$) with $72{,}184$ links and $20.3$M parameters, the rotated schedule $1.721$ ($1.702$, $1.729$, $1.733$) with $74{,}470$ links, and learned coordinates $1.760$ ($1.743$, $1.791$, $1.746$) with $81{,}346$ links.

Two separations survive the seeds and one does not.  The hybrid and the rotated schedule sit $0.057$--$0.060\,\bpc$ below attention with no overlap between seeds; their own difference of $0.003$ is far inside their spread, so we do not rank them against each other.  Learned coordinates sit $0.018$ below attention, but their seeds \emph{overlap} attention's, so on this corpus we claim no separation for them at all.  Every curve is still descending at step $3{,}000$ (Figure~\ref{fig:B2e}, right): this is a comparison at equal step budget, not at convergence, and we have withdrawn ``descends faster'' headlines before (\S\ref{sec:negative}).  What we stand on is the cost at equal quality so far (Table~\ref{tab:B2ecost}): $1/7$ of the links, $42\%$ fewer parameters, and $2.4\times$ less wall-clock time per run on one RTX~5070.

\paragraph{Robustness (Claim B, second half).}  The hybrid stays within $1.718$--$1.801\,\bpc$ over an eight-fold range of learning rates ($0.0005$--$0.004$).  Attention fails to train at $0.002$ and above, three of the seven grid points ($\dagger$ in Table~\ref{tab:B2e}), so its usable window is $0.000125$--$0.001$.  In practice this insensitivity is worth as much as the link count.

\IfFileExists{tables/tabB2e.tex}{
\begin{table}[t]\centering
\caption{Public corpus (enwik8, fingerprint \texttt{c52380b41455}), window 256: held-out bpc over the learning-rate grid, mean of the seeds run at that cell (generated by \texttt{figs.py}; bold = best per schedule; $\dagger$ = failed to train).  \texttt{hy} = hybrid (two attention layers among sixteen), \texttt{spin} = rotated schedule, \texttt{warp} = learned coordinates.}
\label{tab:B2e}
\begin{tabular}{lrrrr}
\toprule
lr & \texttt{attn} & \texttt{hy} & \texttt{spin} & \texttt{warp} \\
\midrule
0.004 & \textdagger & 1.801 & 1.821 & 1.923 \\
0.003 & \textdagger & 1.752 & 1.764 & 1.912 \\
0.002 & \textdagger & 1.718 & \textbf{1.721} & 1.795 \\
0.001 & \textbf{1.778} & \textbf{1.718} & 1.722 & \textbf{1.760} \\
0.0005 & 1.787 & 1.786 & 1.784 & 1.803 \\
0.00025 & 1.881 & 1.924 & 1.925 & 1.977 \\
0.000125 & 2.112 & 2.204 & 2.202 & 2.371 \\
\bottomrule
\end{tabular}

\end{table}
}{}
\IfFileExists{tables/tabB2ecost.tex}{
\begin{table}[t]\centering
\caption{Public corpus, window 256: each schedule at its best learning rate among the cells with three seeds.  \texttt{sd} is the population standard deviation over those seeds; \texttt{links} counts connections per layer, \texttt{weights} all trainable parameters, \texttt{s/run} wall-clock seconds for $3{,}000$ steps on one RTX~5070.  Ratios are relative to attention.}
\label{tab:B2ecost}
\begin{tabular}{lrrrrrrr}
\toprule
form & lr & bpc & sd & seeds & links & weights & s/run \\
\midrule
\texttt{attn} & 0.001 & 1.778 & 0.0090 & 3 & 526336 (1.00$\times$) & 34998473 (1.00$\times$) & 447 (1.00$\times$) \\
\texttt{hy} & 0.002 & 1.718 & 0.0159 & 3 & 72184 (0.14$\times$) & 20318409 (0.58$\times$) & 182 (0.41$\times$) \\
\texttt{spin} & 0.002 & 1.721 & 0.0139 & 3 & 74470 (0.14$\times$) & 20318409 (0.58$\times$) & 312 (0.70$\times$) \\
\texttt{warp} & 0.001 & 1.760 & 0.0219 & 3 & 81346 (0.15$\times$) & 20343009 (0.58$\times$) & 218 (0.49$\times$) \\
\bottomrule
\end{tabular}

\end{table}
\begin{figure}[t]\centering
\includegraphics[width=.48\linewidth]{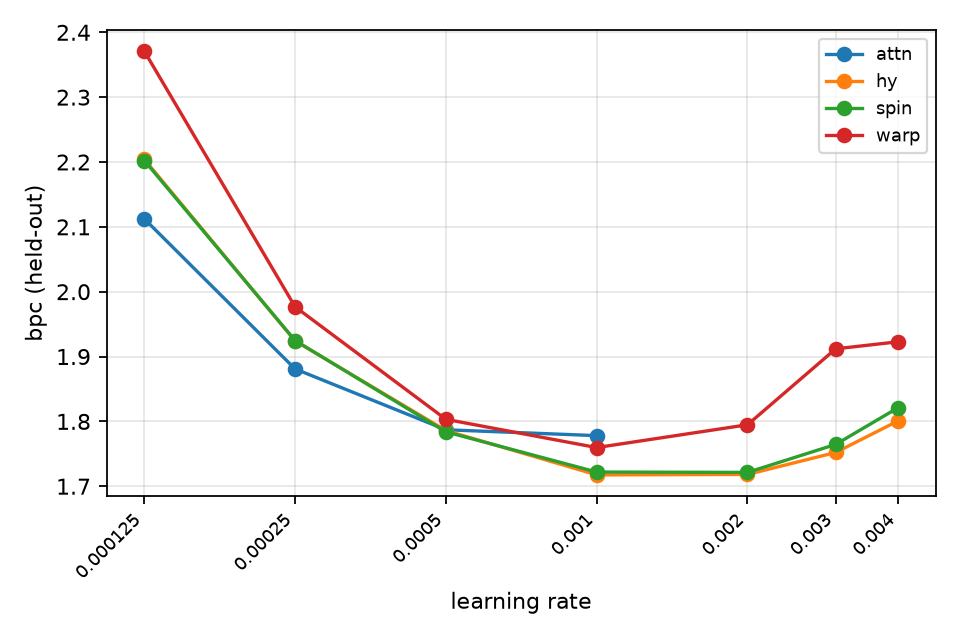}\hfill
\includegraphics[width=.48\linewidth]{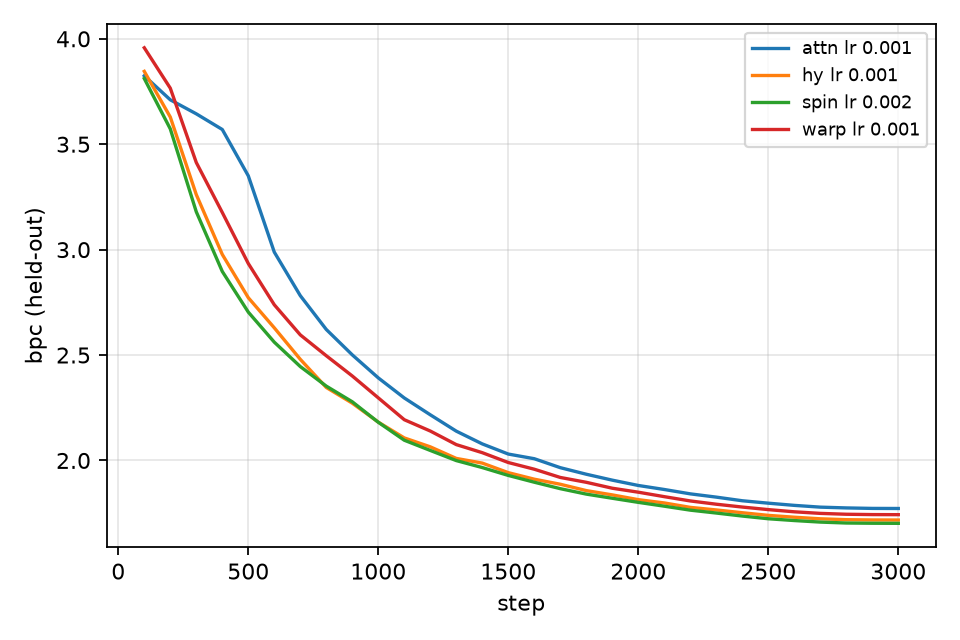}
\caption{Public corpus, window 256: learning-rate grid (left) and held-out curves at each schedule's best rate (right).}
\label{fig:B2e}
\end{figure}
}{}

\subsection{Second distribution: repository corpus, window 256}
\label{sec:w256}
The same grid on the repository corpus (fingerprint \texttt{e1634705154f}).  At its best rate, attention reaches $2.385\,\bpc$ (seeds $2.381$, $2.370$, $2.405$) with $526{,}336$ links and $39.7$M parameters.  The rotated schedule reaches $2.231$ ($2.227$, $2.268$, $2.200$) with $74{,}470$ links and $25.0$M parameters, the \textbf{hybrid} $2.228$ ($2.221$, $2.262$, $2.200$) with $72{,}184$ links and $25.0$M, and learned coordinates $2.248$ ($2.259$, $2.243$, $2.241$) with $81{,}346$ links.
Among the three sparse schedules the gaps of $0.003$--$0.020$ lie inside the seed spread of $0.018$--$0.068$, so we do not rank them.  All three sit $0.14$--$0.16\,\bpc$ below attention with no overlap between seeds.  The parameter counts differ from the public corpus only through the embedding and output layers, because this corpus has $4{,}769$ distinct symbols against enwik8's $201$.

The ordering of the four schedules is the same on both corpora, and the two robustness findings are the same.  What changes is the size of the gap: $0.06\,\bpc$ on the public corpus against $0.16$ here, and learned coordinates separate from attention here but not there.  We read the direction as the transferable part and the magnitude as corpus-specific.

\IfFileExists{tables/tabB2.tex}{
\begin{table}[t]\centering
\caption{Repository corpus, window 256: held-out bpc over the learning-rate grid, mean of the seeds run at that cell (generated by \texttt{figs.py}; bold = best per schedule; $\dagger$ = failed to train).  Tags as in Table~\ref{tab:B2e}.}
\label{tab:B2}
\begin{tabular}{lrrrr}
\toprule
lr & \texttt{attn} & \texttt{hy} & \texttt{spin} & \texttt{warp} \\
\midrule
0.004 & \textdagger & 2.287 & 2.275 & 2.357 \\
0.003 & \textdagger & 2.272 & 2.264 & 2.280 \\
0.002 & \textdagger & \textbf{2.228} & \textbf{2.231} & \textbf{2.248} \\
0.001 & \textbf{2.385} & 2.267 & 2.267 & 2.329 \\
0.0005 & 2.397 & 2.362 & 2.368 & 2.434 \\
0.00025 & 2.599 & 2.581 & 2.584 & 2.691 \\
0.000125 & 2.973 & 2.960 & 2.959 & 3.100 \\
\bottomrule
\end{tabular}

\end{table}
\begin{figure}[t]\centering
\includegraphics[width=.48\linewidth]{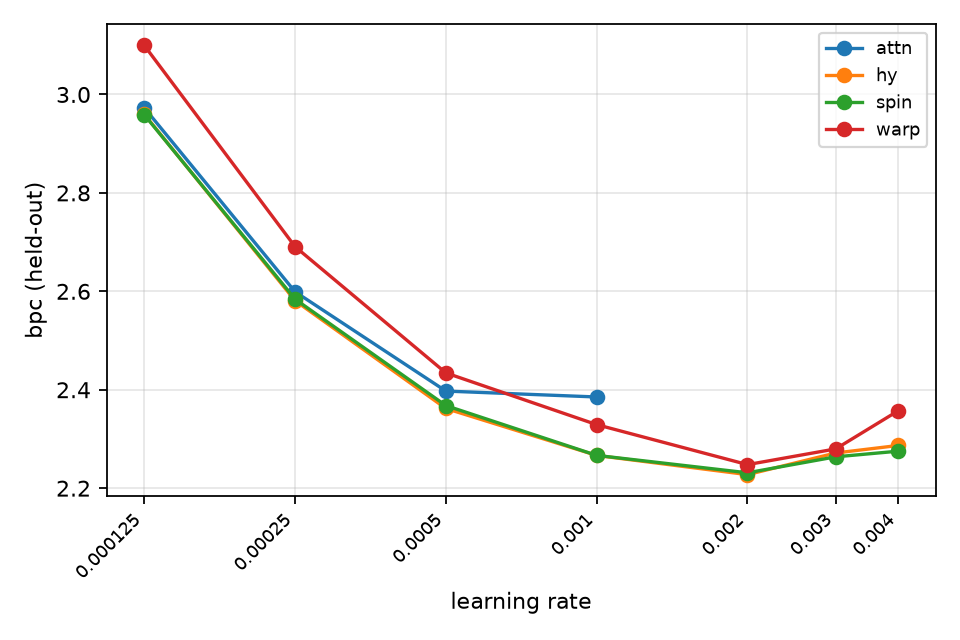}\hfill
\includegraphics[width=.48\linewidth]{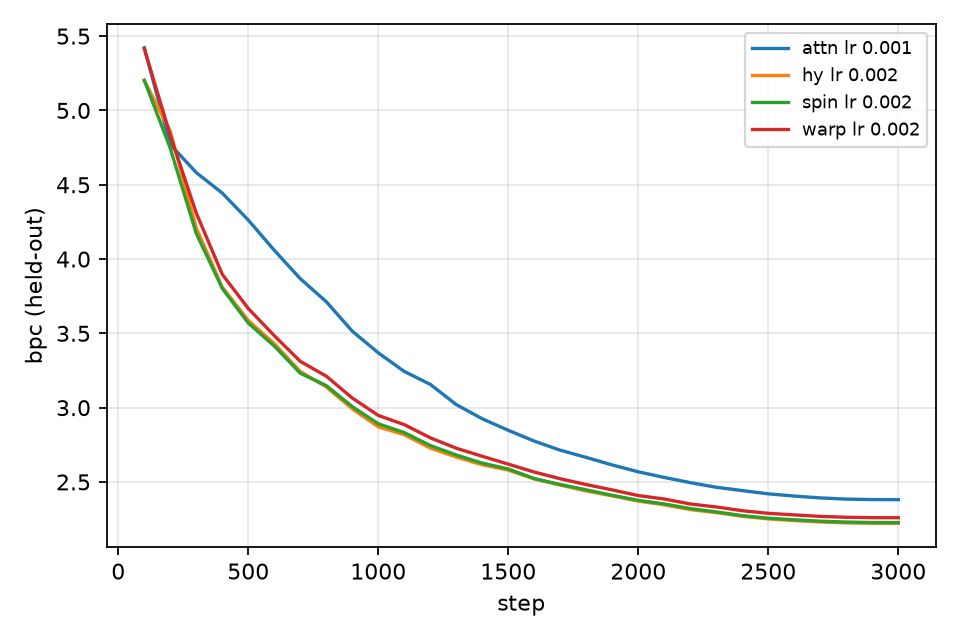}
\caption{Repository corpus, window 256: learning-rate grid (left) and held-out curves at each schedule's best rate (right).}
\label{fig:B2}
\end{figure}
}{}

\subsection{Window 64, 1.7--2.9M parameters: where the cheap schedules stop working}
\label{sec:w64}
Twelve layers, width $128$, $3{,}500$ steps on the repository corpus, three seeds at $\mathrm{lr}=0.002$ for every schedule.  The \textbf{hybrid} reaches $2.796\,\bpc$ (seeds $2.793$, $2.793$, $2.802$) with $5{,}306$ links, level with attention's $2.800$ ($2.797$, $2.804$, $2.800$) at $24{,}960$ links; the seeds overlap, so we claim a match and not a win, at $1/4.7$ of the links and $23\%$ fewer parameters.  Dense mixing, with attention's link count but no search, reaches $2.910$.

The purely sparse schedules do \emph{not} reach dense mixing at this size: one long jump per scale gives $2.943$ at $1/18.7$ of the links, local-only $2.980$, and plain rotation $3.757$.  The gap between the sparse schedules and dense mixing is larger than any seed spread in the table.  This contradicts what we reported before the corpus was frozen, and we withdraw it (\S\ref{sec:negative}).  The reading we are left with is that the hybrid's two attention layers are doing the work that a pure schedule cannot do at width $128$, and that the sparse schedules need the larger model of \S\ref{sec:w256e} before they become competitive.

\IfFileExists{tables/tabB1cost.tex}{
\begin{table}[t]\centering
\caption{Window 64, repository corpus: every schedule at $\mathrm{lr}=0.002$ with three seeds.  Columns as in Table~\ref{tab:B2ecost}.  \texttt{mixa} = local links plus one long jump per scale, \texttt{rot} = plain rotation, \texttt{local} = local links only, \texttt{dense} = all-to-all mixing without search.}
\label{tab:B1cost}
\begin{tabular}{lrrrrrrr}
\toprule
form & lr & bpc & sd & seeds & links & weights & s/run \\
\midrule
\texttt{attn} & 0.002 & 2.800 & 0.0029 & 3 & 24960 (1.00$\times$) & 2866849 (1.00$\times$) & 70 (1.00$\times$) \\
\texttt{local} & 0.002 & 2.980 & 0.0031 & 3 & 1524 (0.06$\times$) & 2080417 (0.73$\times$) & 47 (0.67$\times$) \\
\texttt{rot} & 0.002 & 3.757 & 0.0013 & 3 & 1152 (0.05$\times$) & 2080417 (0.73$\times$) & 46 (0.66$\times$) \\
\texttt{hybrid} & 0.002 & 2.796 & 0.0042 & 3 & 5306 (0.21$\times$) & 2211489 (0.77$\times$) & 87 (1.24$\times$) \\
\texttt{mixa} & 0.002 & 2.943 & 0.0027 & 3 & 1338 (0.05$\times$) & 2080417 (0.73$\times$) & 59 (0.84$\times$) \\
\texttt{dense} & 0.002 & 2.910 & 0.0055 & 3 & 24960 (1.00$\times$) & 2080417 (0.73$\times$) & 47 (0.67$\times$) \\
\bottomrule
\end{tabular}

\end{table}
}{}

\section{Hierarchy changes trainability, not capacity}
\label{sec:hierarchy}

On a grouped associative-recall task (six seeds per condition; three difficulty settings), adding the coarse hypercube never moved the asymptote---$99.3$--$99.9\%$ with and without it in all three settings---but changed how the model gets there:
\begin{center}
\begin{tabular}{lrrr}
\toprule
steps & without & with & seed spread (without / with) \\
\midrule
500  & 68.5\% & 98.9\% & 42.8 / 1.0 points \\
1000 & 97.3\% & 99.7\% & 2.5 / -- \\
2500 & 99.4\% & 99.9\% & 2.5 / -- \\
\bottomrule
\end{tabular}
\end{center}
At $500$ steps the two conditions are completely separated (permutation $p = 0.0011$), and the seed spread differs by $43\times$.  Across the three settings the reached accuracy never moved while speed and spread moved every time (Claim C).  The hierarchy is an optimisation device, not a capacity device.  We have not yet measured it on language.

\section{Dynamics: what a saturated kernel hides}
\label{sec:dynamics}

\paragraph{The artefact.}  Early runs of the dynamics variant (\S\ref{sec:method-dyn}) showed $T=8$ beating $T=1$ on associative recall and the hierarchy helping.  Measuring the effective number of partners (the exponential of the row entropy of the coupling matrix) revealed the mechanism: it was $1.0$ before training and $1.26$ after---the kernel $\exp(-d^2/\tau)$ with $d^2 \approx 100$ and $\tau=1$ underflows to a one-hot, and the model was a chain of pointers, not a field.  Associative recall happens to be solvable by pointer chasing, which is why the numbers were correct and the interpretation was not.

\paragraph{After the fix.}  Dividing $d^2$ by its causal row mean raises the effective partners to $\approx 60$ and the variant behaves as a field.  On language, with the corpus frozen and both learning-rate minima inside the grid:
\IfFileExists{tables/tabC3.tex}{
\begin{table}[h]\centering
\caption{Dynamics variants on language, window 256: held-out bpc over the learning-rate grid, frozen corpus (generated by \texttt{figs.py}).  \texttt{dyn} = unscaled kernel, \texttt{fld} = scaled, \texttt{ph} = scaled with velocity; the digit is $T$.}
\label{tab:C3}
\begin{tabular}{lrrrrrr}
\toprule
lr & \texttt{dyn1} & \texttt{dyn8} & \texttt{fld1} & \texttt{fld8} & \texttt{ph1} & \texttt{ph8} \\
\midrule
0.008 & 2.512 & 3.021 & 2.542 & 2.498 & 2.540 & 2.517 \\
0.004 & \textbf{2.482} & 3.394 & \textbf{2.506} & \textbf{2.461} & \textbf{2.515} & 2.477 \\
0.002 & 2.483 & \textdagger & 2.512 & 2.464 & 2.521 & \textbf{2.469} \\
0.001 & 2.540 & 2.979 & 2.572 & 2.526 & 2.580 & 2.533 \\
0.0005 & --- & \textbf{2.671} & --- & --- & --- & --- \\
\bottomrule
\end{tabular}

\end{table}
}{(Table C3 not generated; run \texttt{figs.py --forms fld1,fld8,ph1,ph8,dyn1,dyn8 --out paper --name C3}.)}
Going from $T=1$ to $T=8$ \emph{hurts} the saturated kernel and \emph{helps} the scaled forms.  The sign reverses on the same corpus and the same grid.  With three seeds at each form's best rate, the scaled field goes from $2.506$ to $2.461$ ($-0.045$, seed spreads $0.021$ and $0.007$), and the scaled field with velocity from $2.515$ to $2.469$ ($-0.046$, spreads $0.015$ and $0.010$).  The saturated kernel goes from $2.482$ to $2.671$, and that is a lower bound on the damage: its $T=8$ optimum sits at the edge of the grid, and one grid step up the three seeds land at $2.647$, $3.243$ and $3.046$.  The saturated kernel also moves its best learning rate down eight-fold when $T$ goes from $1$ to $8$.  The scaled field does not move, and the velocity form moves by one grid step, which is within noise (Claim D).
Between the two scaled forms at $T=8$ the gap is $0.008$ against a seed spread of $0.016$ (median over the six settings run with three seeds); on this corpus we do not rank them.

\paragraph{The same grid on the public corpus.}  Repeating all six settings on enwik8 reproduces the sign reversal, with smaller magnitudes.  The scaled field goes from $1.870$ to $1.835$ ($-0.035$) and the velocity form from $1.870$ to $1.846$ ($-0.024$), while the saturated kernel goes from $1.873$ to $1.975$ ($+0.102$); none of the three pairs overlaps across seeds.  The saturated kernel again moves its best learning rate down, here four-fold.  The one thing that does \emph{not} reproduce is our refusal to rank the two scaled forms: on this corpus the plain field beats the velocity form by $0.011$ with seed spreads of $0.001$ and $0.002$, so the separation is real and it runs \emph{against} the velocity form.  Taken together with the repository corpus, carrying a velocity has now produced no advantage anywhere and a measurable disadvantage once.

\IfFileExists{tables/tabC3e.tex}{
\begin{table}[h]\centering
\caption{Dynamics variants on the public corpus (enwik8), window 256: held-out bpc over the learning-rate grid (generated by \texttt{figs.py}).  Tags as in Table~\ref{tab:C3}.}
\label{tab:C3e}
\begin{tabular}{lrrrrrr}
\toprule
lr & \texttt{dyn1} & \texttt{dyn8} & \texttt{fld1} & \texttt{fld8} & \texttt{ph1} & \texttt{ph8} \\
\midrule
0.008 & 1.881 & \textdagger & 1.877 & 1.845 & 1.880 & 1.858 \\
0.004 & \textbf{1.873} & \textdagger & \textbf{1.870} & \textbf{1.835} & \textbf{1.870} & \textbf{1.846} \\
0.002 & 1.886 & 2.976 & 1.877 & 1.836 & 1.877 & 1.850 \\
0.001 & 1.920 & \textbf{1.975} & 1.923 & 1.883 & 1.923 & 1.893 \\
0.0005 & --- & 2.046 & --- & --- & --- & --- \\
\bottomrule
\end{tabular}

\end{table}
}{}

\section{Toward existing models}
\label{sec:deploy}

\paragraph{Replacing attention layers in Qwen3-0.6B.}  We replace $k$ of the $28$ attention layers by Eq.~\eqref{eq:mixa}, freeze everything else, and fine-tune for $2{,}000$ steps on the frozen corpus; the control replaces nothing and fine-tunes the same $k$ attention layers for the same steps, so the two differ only in the mixing mechanism.

\IfFileExists{tables/tabD1.tex}{
\begin{table}[h]\centering
\caption{Qwen3-0.6B, $2{,}000$ fine-tuning steps: held-out bpc after replacing $k$ of $28$ attention layers, against a control that fine-tunes the same $k$ attention layers unchanged (generated by \texttt{figs.py}).  Penalty = sparse $-$ attention.  The unmodified, un-fine-tuned model is at $3.7505$.}
\label{tab:D1}
\begin{tabular}{rrrrr}
\toprule
layers replaced & sparse & attention & penalty & increment \\
\midrule
2 / 28 & 3.7041 & 3.6665 & \textbf{+0.0376} & --- \\
4 / 28 & 3.6912 & 3.5551 & \textbf{+0.1361} & +0.0985 \\
6 / 28 & 3.7008 & 3.4751 & \textbf{+0.2257} & +0.0896 \\
8 / 28 & 3.7315 & 3.3059 & \textbf{+0.4256} & +0.1999 \\
12 / 28 & 3.8444 & 3.1000 & \textbf{+0.7444} & +0.3188 \\
\bottomrule
\end{tabular}

\end{table}
}{}

Only $k=2$ is close to free ($+0.038\,\bpc$).  Beyond that the penalty grows steeply: $+0.136$, $+0.226$, $+0.426$ and $+0.744$ at $k=4,6,8,12$.  Figure~\ref{fig:D1} shows why.  \emph{The sparse variant is flat in $k$}, staying between $3.69$ and $3.84$ across the whole range, while the attention control improves monotonically from $3.67$ to $3.10$.  The sparse layers are not degrading.  They cannot convert the extra trainable capacity into quality the way attention can.  Against the unmodified model the sparse version is still ahead at $k \le 8$ ($-0.046$ to $-0.019$) and behind at $k=12$ ($+0.094$).  That comparison mixes in the benefit of having seen the corpus at all, which is why we report the control-relative penalty instead.

\IfFileExists{figs/figD1.png}{
\begin{figure}[h]\centering
\includegraphics[width=.62\linewidth]{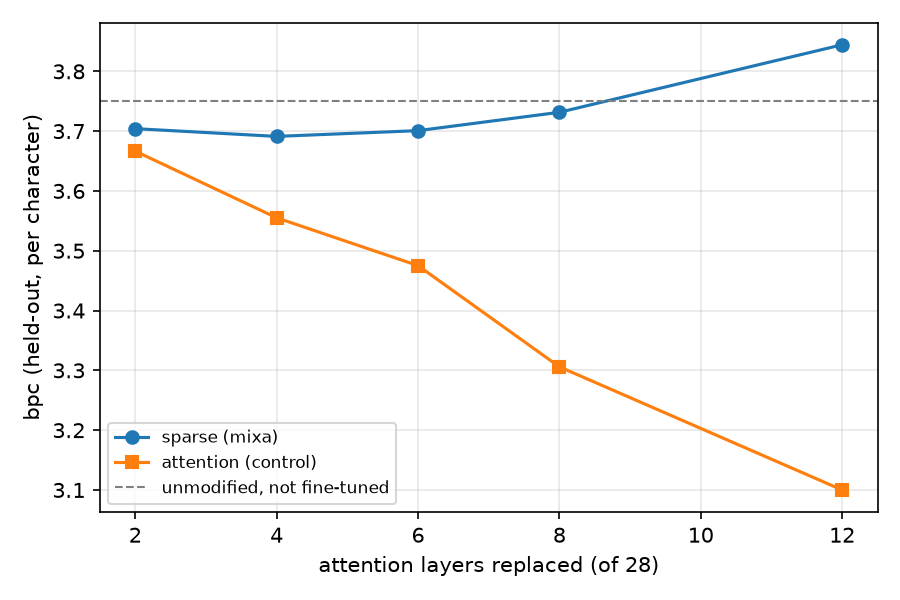}
\caption{Replacing $k$ of $28$ attention layers in Qwen3-0.6B.  The sparse variant is flat; the attention control keeps improving as it is given more layers to fine-tune.}
\label{fig:D1}
\end{figure}
}{}

\paragraph{Why the $500$-step version was optimistic.}  An earlier run at $500$ steps reported $+0.041$, $+0.074$, $+0.119$ and $+0.199$ for $k=2,4,6,8$, two to three times smaller at $k \ge 4$ than the figures above.  At $500$ steps neither side had used its capacity.  The control moved only from $3.692$ to $3.653$ across $k=2$ to $8$, against $3.667$ to $3.306$ at $2{,}000$ steps.  We had predicted the opposite: that a longer budget would \emph{shrink} the penalty, because the replaced layers start from scratch while the control starts from trained weights.  That holds only at $k=2$ ($+0.041 \to +0.038$).  Everywhere else the longer budget favours attention.  The honest reading is that two of $28$ layers is where this substitution is currently free.

\paragraph{Deployability.}  The modified model exports to ONNX using only \texttt{Add, Concat, MatMul, Min, Reshape, Shape, Slice, Squeeze, Sub}, with a maximum deviation of $3.6\times10^{-7}$ from PyTorch, including the one-token-at-a-time decode path (the sparse layer needs only the last $k \le 32$ inputs, so its cache does not grow).  Under 8-bit quantisation the modified model degrades $1.32\times$ as much as the unmodified one (three seeds, no overlap) at $3.2\times$ smaller size.

\section{What did not work, and what we had to fix to know}
\label{sec:negative}

\begin{itemize}
  \item \textbf{A sparse schedule reaching dense mixing at $1/21$ of the links.}  This was our headline for window $64$, and it does not survive the freeze.  Remeasured on the pinned corpus with three seeds, one long jump per scale reaches $2.943\,\bpc$ at $1/18.7$ of the links against dense mixing's $2.910$: it is behind, not level, by more than any seed spread in that table (\S\ref{sec:w64}).  What survives at window $64$ is weaker and different---the \emph{hybrid} draws level with attention at $1/4.7$ of the links.  The pre-freeze runs had been taken on a corpus that changed between them, which is exactly the failure this paper's measurement discipline was built to catch, and it caught our own best number.
  \item \textbf{Learned coordinates.}  Letting positions learn where they sit on the cube (``warp'') buys nothing.  On the repository corpus it reaches $2.248\,\bpc$ against $2.228$ for the fixed hybrid, inside the seed spread, for $13\%$ more links (Table~\ref{tab:B2}).  On the public corpus it is the one schedule whose seeds overlap attention's, so it does not even separate from the baseline there.  A pre-freeze run had shown it \emph{losing}, with four times the seed spread; on frozen data that disappeared too.  We keep only the claim both corpora support: no advantage.
  \item \textbf{Velocity as state.}  No measurable difference from the scaled field on the repository corpus, and on the public corpus a measurable difference in the wrong direction ($0.011\,\bpc$ worse, seed spreads $0.001$ and $0.002$).
  \item \textbf{Transferring the coupling rule to task allocation among software agents}: no effect.
  \item \textbf{Reporting the minimum along the curve.}  Our first window-256 headline ($0.197\,\bpc$ ahead of attention) was the minimum over a $20{,}000$-step run whose floor came at step $2{,}500$, before the learning-rate schedule had begun to anneal; it measured which model \emph{descends faster}, not which is better.  We withdrew it, then withdrew two further versions (``attention needs more steps''; ``attention is fragile'') that were both artefacts of a learning rate tuned for one side only.
  \item \textbf{Evaluation noise.}  Sampling eight random validation batches gave a $0.006$ spread on the same model.  Evaluating every $500$ steps sampled the curve too coarsely to compare minima $0.015$ apart.  Both are fixed: full coverage, every $100$ steps.
  \item \textbf{A moving corpus.}  The training text was gathered from a live repository; adding two files between runs changed the material while the vocabulary count---our only guard---moved by two.  All language numbers before the freeze carry the $\dagger$ mark above and are excluded from the generated tables.
\end{itemize}
We now refuse to print a ranking when the gap between schedules is below the seed spread, and the table generator refuses to emit a table when fingerprints, step counts or evaluation versions are mixed.

\section{Limitations}
\begin{itemize}
  \item Two corpora of $12$M characters each.  One is public but cut and split in our own way, so our numbers are not comparable with published enwik8 results; the other is comparable with nothing, because the repository it is drawn from is private.
  \item Models up to $40$M parameters, three seeds.
  \item A fixed budget of $3{,}000$ steps at which every window-256 curve is still descending.  The language comparison is at equal steps, not at convergence.
  \item The hierarchy has not been measured on language.
  \item The Qwen3 experiment fine-tunes for $2{,}000$ steps on $4$M characters, at one model size and one seed per point, and only on the repository corpus.
  \item No speed measurements on target hardware.  The wall-clock figures are training throughput on one GPU, measured on a machine that was not otherwise idle.
\end{itemize}

\bibliographystyle{plainnat}

\appendix
\section{Schedule strings}
\label{app:glyph}
Schedules are written in the repository's \texttt{.glyph} files as a list of layer types, one CJK character each: \emph{near} (local, $k=1$), \emph{jump}~$k$ (a link of distance $k$ along one cube axis), \emph{attn} (a full attention layer).  Transliterated, the window-256 hybrid (\texttt{hybrid256.glyph}) is
\begin{center}\texttt{near jump128 near attn near jump32 near jump16 near attn near jump4 near jump2 near jump1}\end{center}
(width $512$, hidden $1024$): sixteen layers, local links as the backbone, one jump per scale from $128$ down to $1$, and attention at two of the sixteen.  The rotated schedule (\texttt{spin256.glyph}) replaces the jumps by rotating the cube axis each layer; the learned-coordinate variant (\texttt{warp256.glyph}) lets each position learn where it sits on the cube.  The generated tables refer to these as \texttt{hy}, \texttt{spin} and \texttt{warp}; the same short names are accepted by \texttt{sweep.py --forms}.

\section{Reproduction}

Neither corpus is shipped with the code; both are rebuilt and then checked
against a fingerprint, and \texttt{figs.py} refuses to tabulate records whose
fingerprints disagree.  The public corpus is downloaded from the published
enwik8 archive, verified by SHA-256, read as Latin-1 and cut at $12$M
characters; \texttt{freeze\_corpus.py --enwik8} stops unless the result
fingerprints as \texttt{c52380b41455}.  The second corpus is rebuilt from a
pinned commit of the repository it is drawn from (\texttt{b33e164}), so any
checkout with full history reconstructs the same $12$M characters, fingerprint
\texttt{e1634705154f}; that repository is private, which is why the public
corpus carries the headline.

Everything in \S\ref{sec:w256e} and the public half of \S\ref{sec:dynamics}:
\begin{verbatim}
python freeze_corpus.py --enwik8      # fingerprint must read c52380b41455
python sweep.py --corpus corpus_enwik8.txt --runs enwik8 \
                --forms attn,hy,spin,warp --seeds 1,2
python sweep.py --corpus corpus_enwik8.txt --runs enwik8 \
                --forms fld1,fld8,ph1,ph8,dyn1,dyn8 \
                --lrs 0.008,0.004,0.002,0.001 --seeds 1,2
python figs.py --dir runs/enwik8 --check
python figs.py --dir runs/enwik8 --forms attn,hy,spin,warp \
               --out paper --name B2e --cost --seeded
python figs.py --dir runs/enwik8 --forms fld1,fld8,ph1,ph8,dyn1,dyn8 \
               --out paper --name C3e --cost --seeded
\end{verbatim}

The second corpus (\S\ref{sec:w256}, \S\ref{sec:w64}) needs a checkout of the
private repository:
\begin{verbatim}
python freeze_corpus.py               # fingerprint must read e1634705154f
python sweep.py --forms attn,hy,spin,warp --seeds 1,2
python sweep.py --forms fld1,fld8,ph1,ph8,dyn1,dyn8 \
                --lrs 0.008,0.004,0.002,0.001 --seeds 1,2
python sweep.py --forms dyn8 --lrs 0.008,0.004,0.002,0.001,0.0005
python figs.py --check
python figs.py --forms attn,hy,spin,warp --out paper --name B2
python figs.py --forms fld1,fld8,ph1,ph8,dyn1,dyn8 --out paper --name C3
python sweep.py --runs w64 --n 64 --L 12 --d 128 --hid 256 \
                --steps 3500 --lrs 0.002 --seeds 1,2
python figs.py --dir runs/w64 --out paper --name B1 --cost --seeded
\end{verbatim}

\end{document}